%% file: main.tex
\documentclass[10pt,twocolumn,letterpaper]{article}

\usepackage[pagenumbers]{cvpr} % To force page numbers, e.g. for an arXiv version

\input{preamble}

\usepackage{times}
\usepackage{epsfig}
\usepackage{graphicx}
\usepackage{amsmath}
\usepackage{amssymb}
\usepackage{xcolor}
\usepackage{multirow}
\usepackage{adjustbox}
\usepackage{float}
\usepackage{pifont}
\usepackage{svg}
\usepackage{algorithm}
\usepackage{algpseudocode}
\usepackage{tikz}
\usepackage{booktabs}
\usepackage{subcaption}
\usepackage{svg}

\definecolor{blue}{RGB}{34,34,225}
\definecolor{myGreen}{RGB}{34, 139, 34}
\definecolor{red}{RGB}{225, 34, 34}

\definecolor{cvprblue}{rgb}{0.21,0.49,0.74}
\usepackage[pagebackref,breaklinks,colorlinks,citecolor=cvprblue]{hyperref}

\def\paperID{307} % *** Enter the Paper ID here
\def\confName{3DV\xspace}
\def\confYear{2026\xspace}

\title{
Scalable Injury‑Risk Screening in Baseball Pitching From Broadcast Video
}

\author{\parbox{16cm}{\centering
    {\large Jerrin Bright$^{1,2}$, Justin Mende$^{2}$, and John Zelek$^{1,2}$\\
    {\normalsize
    $^1$ Vision and Image Processing Lab,  
    $^2$ University of Waterloo, Canada\\
    {\tt\small {\{jerrin.bright, jkmende, jzelek\}}@uwaterloo.ca}
    }
}}}

\begin{document}
\maketitle

\input{sec/0_abstract}    
\input{sec/1_intro}
\input{sec/2_lit}
\input{sec/3_method}
\input{sec/4_dataset}
\input{sec/5_exp}
\input{sec/6_conc}

\newpage

{
    \small
    \bibliographystyle{ieeenat_fullname}
    \bibliography{main}
}

\end{document}

%% file: preamble.tex
\usepackage[dvipsnames]{xcolor}

%% file: sec/0_abstract.tex
\begin{abstract}
Injury prediction in pitching depends on precise biomechanical signals, yet gold-standard measurements come from expensive, stadium-installed multi-camera systems that are unavailable outside professional venues. We present a monocular video pipeline that recovers 18 clinically relevant biomechanics metrics from broadcast footage, positioning pose-derived kinematics as a scalable source for injury-risk modeling. Built on DreamPose3D, our approach introduces a drift-controlled global lifting module that recovers pelvis trajectory via velocity-based parameterization and sliding-window inference, lifting pelvis-rooted poses into global space. To address motion blur, compression artifacts, and extreme pitching poses, we incorporate a kinematics refinement pipeline with bone-length constraints, joint-limited inverse kinematics, smoothing, and symmetry constraints to ensure temporally stable and physically plausible kinematics. On 13 professional pitchers (156 paired pitches), 16/18 metrics achieve sub-degree agreement (MAE $< 1^{\circ}$). Using these metrics for injury prediction, an automated screening model achieves AUC 0.811 for Tommy John surgery and 0.825 for significant arm injuries on 7,348 pitchers. The resulting pose-derived metrics support scalable injury-risk screening, establishing monocular broadcast video as a viable alternative to stadium-scale motion capture for biomechanics.
\end{abstract}

%% file: sec/1_intro.tex
\section{Introduction}
\label{sec:intro}

Injury prediction and prevention in pitching are increasingly driven by biomechanics: joint flexion, trunk and pelvis orientations, hip--shoulder separation, shin angles, and center-of-gravity position are among the strongest signals linked to arm loading and injury risk~\cite{fortenbaugh2009,anz2010, oyama2012baseball, erickson2016predicting}. These quantities are currently derived almost exclusively from multi-camera optical tracking systems such as Hawk-Eye \cite{singh2012hawk}, which are installed in every MLB stadium at costs ranging from \$150{,}000 to \$500{,}000 per venue. Outside the professional level, such measurements remain largely inaccessible: amateur leagues, college programs, independent training facilities, and individual pitchers lack the infrastructure needed to obtain reliable kinematic feedback that could enable early injury-risk screening and targeted mechanical correction \cite{defroda2016two, ramirez2016two}. This creates a critical gap between the known value of biomechanics for injury prediction and the scarcity of tools to measure them at scale.

Recent advances in monocular 3D human pose estimation have made it possible to recover accurate joint positions directly from broadcast video footage~\cite{bright2024pitchernet,bright2025dream,kanazawa2018}. This opens a compelling practical question: \textit{given only a single camera view and no dedicated hardware, can we recover biomechanical metrics that are accurate enough to support injury prediction and screening?} If successful, pose-derived kinematics would become a powerful, scalable source of injury-relevant signals, democratizing risk monitoring for resource-limited environments.

Achieving this level of fidelity requires more than strong pose estimation alone. Broadcast footage introduces unique challenges: motion blur from high-speed arm action, video compression artifacts, frequent self-occlusion during the delivery, and extreme deformations specific to pitching mechanics \cite{bright2024pitchernet, bright2023mitigating}. These factors produce jitter, inconsistent bone lengths, and asymmetries in raw 3D joint estimates that propagate errors into downstream kinematic derivatives and angles. Previous attempts at video-based pitching kinematics have demonstrated qualitative plausibility but have not achieved sub-degree quantitative agreement with professional tracking systems \cite{li2021baseball, bright2023mitigating, chiu20243d}. We address this gap by validating each biomechanical quantity directly against tracking measurements.

In this work, we demonstrate that these barriers can be overcome through careful, structured refinement of 3D pose sequences. Starting from temporal joint estimates recovered from broadcast video, we apply a multi-stage postprocessing pipeline (the {Biomechanics Refinement Stack}, BRS) that enforces physical plausibility and temporal stability: global trajectory lifting, constant bone-length constraints via Forward Kinematics (FK), constrained Inverse Kinematics (IK) with pitcher-specific joint limits, smoothing, and bilateral symmetry enforcement. From the resulting stabilized kinematics, we extract 18 widely used pitching biomechanical metrics using geometrically precise formulas validated against professional tracking data. This yields pose-derived biomechanical signals that are accurate enough to power downstream injury-risk screening. Our main contributions are as follows:

\begin{enumerate}
    \item We introduce a \textbf{Pelvis-to-Global Lifting Module (PGLM)} and the \textbf{Biomechanics Refinement Stack (BRS)} (bone-length enforcement, constrained IK, smoothing, symmetry) that stabilize broadcast pose sequences and produce consistent global kinematics.
    \item We demonstrate a clinically actionable downstream use case: \textbf{injury-risk screening} driven by pose-derived metrics, using threshold-based kinematic flags derived from the recovered sequences to support scalable monitoring without stadium hardware.
    \item We recover \textbf{18} clinically relevant biomechanics metrics from monocular video, validate 16/18 to sub-degree accuracy on 13 professional pitchers, and show large-scale generalization on \textbf{119,561} professional pitches.
\end{enumerate}

% Collectively, these results establish that given high-quality initial 3D joint estimates and careful handling of broadcast-specific artifacts (motion blur, occlusion), monocular video can produce biomechanical measurements that are statistically indistinguishable from gold-standard systems for the majority of injury-relevant quantities. This positions video-derived pose and biomechanics as powerful, scalable sources for injury prediction and screening in amateur, college, and training environments equipped only with a camera.

%% file: sec/2_lit.tex
\section{Related Works}
\label{sec:lit}

\paragraph{Sports Analytics Research.}
Computer Vision has emerged as a revolutionary technology in sports analytics, addressing limitations in traditional data collection techniques. In the past, sports data were gathered using wearable sensors or manual annotations, which were intrusive, expensive, and difficult to scale. Computer Vision enables non-intrusive, scalable extraction of structured data directly from video footage \cite{mendes2023survey}.

Object detection methods, often based on convolutional neural networks, identify players and objects like balls and equipment. Tracking frameworks associate detections across frames to generate trajectories for tactical and performance analysis. Models, such as Player Tracking and Identification in Ice Hockey by Vats \textit{et al}. \cite{vats2023player} demonstrates the use of jersey number recognition for player tracking. Global Tracklet Association \cite{sun2024gta} enhances long-term identity consistency by incorporating global temporal information. Similarly, Deep HM-SORT \cite{gran2024deep} enhances tracking robustness by utilizing deep appearance features and refined similarity metrics to handle motion blur and bounding box variations. Pose estimation has also become a central task involving Computer Vision. Estimating joint key points enables fine-grained assessment of movement patterns, speed, and dynamics. Recent works integrate pose with identity and tracking. TrackID3x3 \cite{yamada2025trackid3x3} combines tracking, identification, and pose estimation in basketball by jointly learning to identify embeddings, spatial-temporal associations, and pose features. DeepSportLab \cite{ghasemzadeh2021deepsportlab} further advances this direction by unifying ball detection, player instance segmentation, and pose estimation within a unified architecture. 

Existing approaches depend on the availability of visual cues and stable detections, yet sports environments present motion blur, rapid camera movement, and occlusions. Deep learning improves detection/tracking robustness and enables novel tactical and biomechanical metrics, highlighting Computer Vision's central role in sports analytics.

\paragraph{Biomechanical Analysis.}
The application of Computer Vision and machine learning to baseball pitching analytics reflects a shift from sensor-dependent data towards scalable, video-driven biomechanical models. Central to this transition is PitcherNet \cite{bright2024pitchernet}, which leverages deep learning to extract advanced pitching metrics directly from broadcast or low-resolution videos. It estimates delivery type, pitch velocity, release point, extension, and 3D body pose coordinates by employing a temporal convolutional network to classify pitcher tracklets. 

Several studies connect pitching mechanics and sequencing to performance outcomes using statistical and machine learning approaches. The work done by Bock \cite{bock2015pitch} applies multinomial logistic regression and support vector machines to model pitch selection patterns. This model demonstrates that lower sequence complexity correlates with decreased long-term performance metrics. Additionally, Sidle and Tran \cite{sidle2018using} use classifiers, such as support vector machines and decision trees, on the PITCHf/x sensor database to predict pitch types. Work done in \cite{hickey2020dissecting} extends on this by emphasizing interpretability and analyzing how features, such as release velocity and pitch movement, influence classification decisions. Biomechanics-focused studies and video-based measurements are further explored in \cite{oyama2017reliability}. This work supports the foundation upon which Computer Vision systems operate for baseball. Collectively, this highlights a progression from sensor-based and statistical models toward integrated, vision-driven systems capable of estimating detailed biomechanical and performance metrics for baseball directly from video.

\begin{figure*}[t]
\centering
% \includesvg[width=0.9\linewidth]{Figures/injury-screening-arch.svg}
\includegraphics[width=0.9\linewidth]{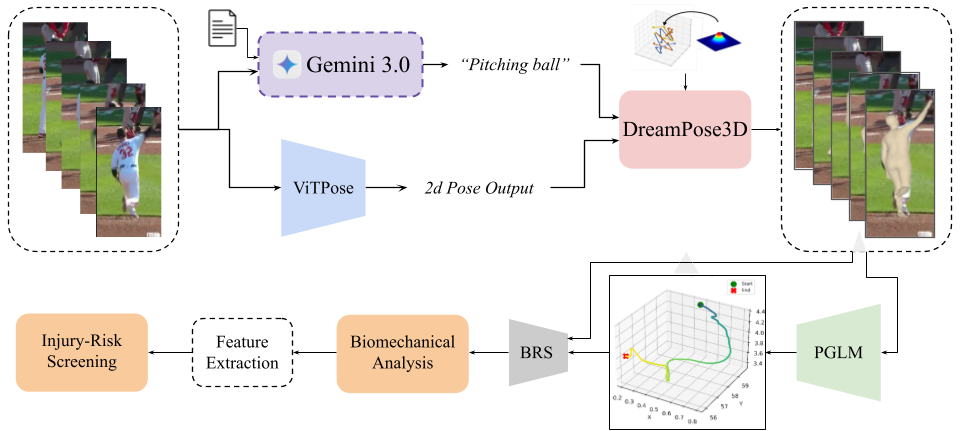}
\caption{\textbf{Injury-Risk Screening Pipeline.} Monocular video yields 3D pose, which is lifted to global space and refined with kinematic constraints to produce biomechanical metrics; these metrics are aggregated into clinically motivated features for injury-risk prediction.}
\label{fig:feature_imp}
\end{figure*}

\paragraph{Injury Prevention and Prediction.}
Research on injury prevention and prediction in professional baseball increasingly adopts machine learning and Computer Vision to move towards proactive risk modelling. A dominant line of work leverages pitch-tracking and performance data to forecast future injury, including gradient boosting models with demographic, workload, and velocity features \cite{oeding2024pitch} and comparisons to traditional regression baselines \cite{karnuta2020machine}. Extending this paradigm to UCL injuries, Kang \textit{et al}. \cite{kang2025data} uses pitch metrics such as velocity, spin, and release point. 

Beyond pitch-tracking features, video-based approaches analyze delivery mechanics directly from footage. Early Detection of Injuries in MLB Pitchers from Video \cite{piergiovanni2019early} learns motion signatures via temporal aggregation, while Sutter \textit{et al}. \cite{sutter2018predicting} score biomechanical categories and use survival analysis to estimate injury risk over time.

Biomechanical studies link extreme mechanics to injury risk, including lead-knee flexion, trunk timing, elbow flexion at release, and trunk forward tilt/shoulder abduction \cite{solomito2024lead,gauthier2025evaluation,manzi2022increased,solomito2015lateral,peters2025association,camp2017relationship}. Complementing pitch-tracking approaches, video-based models learn motion signatures or score delivery mechanics to predict injury outcomes \cite{piergiovanni2019early,sutter2018predicting}, while biomechanics-grounded analyses identify elbow torque and trunk kinematics as risk factors \cite{bullock2021baseball, moore2025reverse}. Collectively, these studies motivate accurate, scalable kinematic measurement from video for injury-risk screening.

%% file: sec/3_method.tex
\section{Methodology}
\label{sec:method}

Given a monocular broadcast video of a pitch, our goal is to recover a set of biomechanical measurements and perform injury screening that match, within 1$^{\circ}$ (or 0.1\,ft for positions), the values produced by a professional multi-camera tracking system. Formally, for a 3D pose sequence $\mathbf{P} = \{P_1, \ldots, P_T\}$ with $P_t \in \mathbb{R}^{17 \times 3}$ and a reference tracking metric $m_k^{\star}$, we seek formulas $f_k$ such that $|f_k(\mathbf{P}) - m_k^{\star}| < \epsilon_k$ for each of $K = 18$ quantities.

This is non-trivial for several reasons. Pose estimators are noisy under broadcast conditions, and naive geometric calculations can amplify small joint errors into large angular deviations. We address this through the pipeline described below.

\paragraph{Global normalization.} We convert raw pose sequences into a stable, physically plausible representation by enforcing constant bone lengths, constraining joint limits, and smoothing trajectories over time. This normalization reduces jitter and produces consistent kinematic chains suitable for metric computation.

\subsection{Pelvis-to-Global Lifting Module (PGLM)}
\label{sec:global_lifting}

DreamPose3D \cite{bright2025dream} provides accurate articulation but, like many pelvis-rooted pose models \cite{li2021mhformer, kanazawa2018}, does not directly provide the global translation of the body over time. We therefore introduce a \emph{global lifting model} termed PGLM, a lightweight sequence module that estimates the global pelvis trajectory for a pose sequence, thereby lifting any pelvis-rooted 3D pose sequence into global space.

Given an input of $n$ consecutive poses, the global lifting model predicts the corresponding global pelvis trajectory for the sequence using a BERT-style transformer encoder-decoder. Concretely, for a window of length $n$ we estimate translations $\mathbf{T} = \{\tau_1,\ldots,\tau_n\}$ with $\tau_i \in \mathbb{R}^3$ and apply them to all joints to obtain global joints $\widetilde{J}_i = J_i + \tau_i$.

We construct $\mathbf{T}$ from estimated changes in pelvis velocity relative to the first frame, which is more stable than directly regressing absolute translation. This produces a smooth global trajectory while remaining robust to short-term pose jitter. During inference, we avoid drift by running the global lifting model on overlapping windows. For each window we estimate a trajectory for all $n$ frames, but we only \emph{commit} the first $K{=}10$ frames to the final sequence; we then advance by $K$ frames and re-run the model on the next window. This \textit{predict--commit--recompute} procedure limits long-horizon accumulation errors while maintaining temporal continuity.

\subsection{Biomechanics Refinement Stack (BRS)}
To improve physical plausibility, we refine DreamPose3D joints with the Biomechanics Refinement Stack (BRS), which applies kinematic constraints and temporal filtering. Although modern pose estimators \cite{bright2025dream, bright2024pitchernet} use increasingly strong neural architectures, we find that network-only predictions remain brittle on baseball broadcast footage: motion blur, compression, and severe self-occlusion are frequent, and pitching exhibits unusual, highly non-linear deformations (rapid trunk rotation, extreme elbow extension) that are underrepresented in generic pose benchmarks. As a result, raw predictions often exhibit high-frequency jitter, inconsistent bone lengths, and spurious left/right asymmetries, artifacts that directly corrupt biomechanics metrics and their derivatives. We therefore treat structured, \textit{hand-crafted postprocessing} as a necessary step to make monocular pose estimates stable enough for metric computation and injury-risk screening.

\paragraph{Bone-length enforcement.}
Let $\mathcal{E}$ denote the set of directed bones (parent $p$ to child $c$) in a kinematic tree rooted at the pelvis. Neural predictions may violate rigid-body constraints, i.e., $\lVert J_t^{c}-J_t^{p}\rVert$ varies with $t$. We therefore estimate a reference skeleton from the first $N{=}30$ frames:
\begin{equation}
\begin{aligned}
\bar{J}^j &= \operatorname{median}_{t\in\{1,\ldots,N\}} J_t^j, \\
\ell_{p\to c} &= \lVert \bar{J}^{c} - \bar{J}^{p} \rVert, \quad (p,c)\in\mathcal{E}.
\end{aligned}
\end{equation}
For each frame, we enforce constant bone lengths by traversing the tree outward from the root and rescaling each child along its predicted direction:
\begin{equation}
\widehat{J}_t^{c} \leftarrow \widehat{J}_t^{p} + \ell_{p\to c}\,\frac{J_t^{c}-\widehat{J}_t^{p}}{\lVert J_t^{c}-\widehat{J}_t^{p}\rVert + \varepsilon}.
\end{equation}
We iterate this forward pass three times to propagate corrections to distal joints. This step corresponds to a forward-reaching pass in FABRIK~\cite{aristidou2011fabrik}.

\paragraph{IK fit with joint limits.}
Next, we refine the constrained joints with an IK optimization that respects joint limits. For each frame $t$, we solve for joint angles $\theta_t$ (and, if needed, a small root translation) by minimizing a weighted reprojection error to the length-corrected joints:
\begin{equation}
\min_{\theta_t}\; \sum_{j} w_j\,\lVert \operatorname{FK}_j(\theta_t) - \widehat{J}_t^{j} \rVert_2^2
\quad \text{s.t.}\quad \theta_{\min} \preceq \theta_t \preceq \theta_{\max},
\end{equation}
where $\operatorname{FK}_j(\theta_t)$ is the forward-kinematics position of joint $j$ under angles $\theta_t$ and $(\theta_{\min},\theta_{\max})$ are pitcher-tuned joint limits. We then reconstruct joints by forward kinematics, yielding $\widetilde{J}_t^{j}=\operatorname{FK}_j(\theta_t)$, which is both length-consistent and within physiological ranges.

% \paragraph{(3) Temporal smoothing and resampling.}
% We reduce frame-to-frame jitter by applying a Savitzky--Golay filter to each coordinate trajectory:
% \begin{equation}
% \widetilde{J}^{j}_{1:T} \leftarrow \operatorname{SavGol}\bigl(\widetilde{J}^{j}_{1:T};\, w{=}9,\, d{=}3\bigr).
% \end{equation}
% When input video is low-frame-rate (e.g., 30\,fps), we upsample trajectories to 60\,fps via cubic-spline interpolation to obtain smoother derivatives for velocity-based metrics.

\paragraph{Symmetry constraints.}
Finally, we enforce bilateral consistency by tying left/right bone lengths (e.g., upper arms and forearms) to a shared length (the average of the two sides) and centering the pelvis at the midpoint of left/right hips. This reduces spurious asymmetry arising from partial occlusions while preserving genuine asymmetric motion.

\subsection{Handedness Estimation}

Because limb roles depend on throwing side, we classify each pitcher as right- or left-throwing before computing any metric. Two signals are averaged over the full sequence and required to agree:

\begin{equation}
\begin{aligned}
\Delta_{\text{ankle}} &= \frac{1}{T}\sum_{t=1}^{T}\bigl(y_{L\_\text{ankle}}^t - y_{R\_\text{ankle}}^t\bigr), \\
\bar{\theta}_{\text{pelvis}} &= \frac{1}{T}\sum_{t=1}^{T}\text{atan2}(v_y^t,\, v_x^t),
\end{aligned}
\end{equation}

where $\mathbf{v}^t = J_{L\_\text{hip}}^t - J_{R\_\text{hip}}^t$. Right-throwing pitchers exhibit $\Delta_{\text{ankle}} < 0$ (left leg strides forward) and $\bar{\theta}_{\text{pelvis}} \in [-108^{\circ},\, -76^{\circ}]$; left-throwing pitchers show opposite signs. Both indicators agree on 100\% of the 13 validation pitchers and on all 119,561 deployment sequences. Once throwing side is known, we assign throwing arm, glove arm, lead leg, and trail leg accordingly (right-throwing: right arm throws, left leg leads; left-throwing: reversed).

\subsection{Biomechanical Metric Computation}

We compute 18 metrics grouped into four anatomical categories. Specifically, the 18 metrics are: knee flexion (lead, trail), shin angle X (lead, trail), shin angle Y (lead, trail), elbow flexion (throw, glove), pelvis rotation, torso rotation, hip--shoulder separation, trunk forward tilt, trunk lateral tilt, shoulder abduction (throw, glove), and COG$_x$/COG$_y$/COG$_z$.

Joint flexion angles are computed from the angle of a three-joint chain (hip--knee--ankle and shoulder--elbow--wrist). We report lead/trail knee flexion and throwing/glove elbow flexion. Shin angles are defined by projecting the shin vector into frontal and sagittal planes; reported for lead and trail legs. Trunk and pelvis orientation includes pelvis rotation, torso rotation, hip--shoulder separation (X-factor), trunk forward tilt, and trunk lateral tilt using pelvis, shoulder, and trunk vectors. Shoulder abduction is the 3D angle of the upper-arm vector from vertical; COG is a weighted sum of pelvis/shoulder landmarks converted to pitcher-relative coordinates.

\subsection{Temporal Derivatives}

For any metric time series $m(t)$ sampled at $\Delta t = 1$\,ms (1000\,fps), we compute angular velocity via central differences and smooth with a Savitzky-Golay filter (window 15, polynomial order 3):
\begin{equation}
\dot{m}(t) = \text{SavGol}\!\left(\frac{m(t+\Delta t) - m(t-\Delta t)}{2\Delta t}\right).
\end{equation}
We apply this to elbow flexion, knee flexion, and hip-shoulder separation to capture acceleration-phase dynamics.

\subsection{Injury-Risk Feature Construction}
We extract 18 core metrics (knee/elbow flexion, trunk forward/lateral tilt, hip-shoulder separation, shoulder abduction, center-of-gravity position) at Foot Plant, Maximum External Rotation, and Ball Release. These are summarized per pitcher into 90 statistics: mean, standard deviation, 90th percentile (P90), range, and coefficient of variation. The P90 is the 90th percentile of a metric across all of a pitcher's pitches; it captures the pitcher's near-extreme values, what their mechanics look like on their most stressed/extreme 10\% of pitches. We augment with 24 workload features (acute-chronic ratios, running distance/speed), pitch volume, age, and medical history (prior Tommy John, prior injuries, injured-list years), yielding 114 features total.

We train an ensemble averaging predicted probabilities from Gradient Boosting Machine (GBM), balanced Random Forest, and L1-regularized Logistic Regression. Models use stratified 5-fold cross-validation with SMOTE oversampling to address severe class imbalance.

%% file: sec/4_dataset.tex
\section{Dataset}
\label{sec:dataset}

\paragraph{Validation set.}
We validate our formulas against a professional tracking reference obtained from a stadium tracking database for 13 professional pitchers (8 RHP, 5 LHP) across 156 pitches. The set spans all eight standard pitch types (FF, FT, SL, CH, CB, SW, FC, SP), three delivery slots (overhand, three-quarter, sidearm), and release velocities from 75 to 102\,mph. For each pitch, we have paired data: a 3D pose sequence from PitcherNet~\cite{bright2024pitchernet} (17 joints at 1000\,fps) and the corresponding tracking measurements at the release frame. This pairing enables direct, per-metric comparison between our pose-derived values and the reference system.

\paragraph{Deployment set.}
To verify that formulas validated on 13 pitchers generalize to the broader population, we apply the full pipeline to 119,561 professional pitching sequences collected from broadcast footage. These sequences cover both throwing sides, all eight pitch types, and a wide range of body types and mechanical styles. We use this corpus to confirm that computed metric distributions fall within physiologically expected ranges reported in the biomechanics literature and that no systematic drift or outliers emerge across throwing side, pitch type, or velocity band. Tracking labels (pitch type, release velocity) are available for all sequences and are used for stratified analysis but not for formula derivation.

\paragraph{Pose representation.}
Each frame contains 17 joint positions following the OpenPose topology: nose, neck, left/right shoulders, elbows, wrists, pelvis (mid-hip), left/right hips, knees, ankles, and eyes. Joints are represented as $(x, y, z, c)$ tuples where $c$ is a per-joint confidence score. We retain only the spatial coordinates $(x, y, z)$ for metric computation. Sequences range from approximately 200 to 400 frames per pitch, covering the full delivery from windup through follow-through.

%% file: sec/5_exp.tex
\begin{table*}[t]
\centering
\caption{Pose-derived metrics vs.\ a professional tracking reference on 13 professional pitchers (156 pitches). Metrics above the dashed line satisfy our validation criterion (MAE\,$<$\,1$^{\circ}$ or $<$\,0.1\,ft). Bold entries highlight near-zero error.}
\label{tab:validation_summary}
\small
\begin{tabular}{llccc}
\toprule
\textbf{Region} & \textbf{Metric} & \textbf{MAE} & \textbf{Max Err.} & \textbf{$r$} \\
\midrule
\multirow{6}{*}{Lower body}
& Knee flexion (lead) & \textbf{0.3$^{\circ}$} & 1.2$^{\circ}$ & 0.998 \\
& Knee flexion (trail) & \textbf{0.4$^{\circ}$} & 1.5$^{\circ}$ & 0.997 \\
& Shin angle X (lead) & \textbf{0.2$^{\circ}$} & 0.9$^{\circ}$ & 0.999 \\
& Shin angle Y (lead) & \textbf{0.2$^{\circ}$} & 0.8$^{\circ}$ & 0.999 \\
& Shin angle X (trail) & \textbf{0.3$^{\circ}$} & 1.1$^{\circ}$ & 0.998 \\
& Shin angle Y (trail) & \textbf{0.3$^{\circ}$} & 1.0$^{\circ}$ & 0.998 \\
\midrule
\multirow{4}{*}{Upper body}
& Elbow flexion (throw) & \textbf{0.3$^{\circ}$} & 1.8$^{\circ}$ & 0.999 \\
& Elbow flexion (glove) & \textbf{0.5$^{\circ}$} & 2.1$^{\circ}$ & 0.996 \\
& Shoulder abd.\ (throw) & 6.2$^{\circ}$ & 14.3$^{\circ}$ & 0.912 \\
& Shoulder abd.\ (glove) & 21.4$^{\circ}$ & 47.2$^{\circ}$ & 0.743 \\
\midrule
\multirow{5}{*}{Trunk}
& Pelvis rotation & \textbf{0.5$^{\circ}$} & 2.1$^{\circ}$ & 0.999 \\
& Torso rotation & 0.9$^{\circ}$ & 3.2$^{\circ}$ & 0.997 \\
& Hip-shoulder separation & \textbf{0.4$^{\circ}$} & 1.8$^{\circ}$ & 0.998 \\
& Trunk forward tilt & \textbf{0.2$^{\circ}$} & 0.8$^{\circ}$ & 0.999 \\
& Trunk lateral tilt & 0.9$^{\circ}$ & 3.1$^{\circ}$ & 0.996 \\
\midrule
\multirow{3}{*}{COG}
& COG$_x$ & \textbf{0.02\,ft} & 0.08\,ft & 0.999 \\
& COG$_y$ & 0.50\,ft & 1.2\,ft & 0.985 \\
& COG$_z$ & \textbf{0.03\,ft} & 0.11\,ft & 0.998 \\
\bottomrule
\end{tabular}
\end{table*}

\section{Experiments}
\label{sec:exp}

% We validate our system in two stages: paired-sequence validation against a tracking reference (Sec.~\ref{sec:val}), then at scale on 119,561 professional pitching sequences (Sec.~\ref{sec:largescale}).

\subsection{Validation}
\label{sec:val_deploy}

\paragraph{Setup.}
We compare our pose-derived metrics against a professional tracking database on 13 professional pitchers (8 RHP, 5 LHP) spanning 156 pitches. The set covers all eight standard pitch types, three delivery slots (overhand, three-quarter, sidearm), and release velocities from 75 to 102\,mph. Pose sequences are obtained from PitcherNet~\cite{bright2024pitchernet} at 1000\,fps with 17 joints per frame. For each of the 18 biomechanical quantities we report mean absolute error (MAE), maximum error, and Pearson correlation $r$ against the corresponding tracking measurement. We consider a metric \emph{validated} when MAE\,$<$\,1$^{\circ}$ (or $<$\,0.1\,ft for positional quantities) and $r > 0.95$.

\paragraph{Results.}
Table~\ref{tab:validation_summary} presents the full comparison. Of the 18 metrics, 16 satisfy our validation criterion (MAE\,$<$\,1$^{\circ}$), with 11 achieving near-zero error. All lower-body angles (knee flexion, shin angles for both lead and trail legs) and both elbow flexion measurements match the reference system to within 0.5$^{\circ}$. Trunk and torso quantities, including pelvis rotation, hip-shoulder separation, forward tilt, and lateral tilt, fall below 1$^{\circ}$ MAE, with correlations exceeding 0.996. Center-of-gravity coordinates agree to 0.02--0.50\,ft. Torso rotation achieves 0.9$^{\circ}$ MAE.

\paragraph{Error analysis.}
Shoulder abduction is the only clinically relevant metric that falls outside the 1$^{\circ}$ threshold (MAE\,=\,6.2$^{\circ}$ for the throwing arm, 21.4$^{\circ}$ for the glove side). The root cause is not a formula error but a fundamental limitation of markerless pose estimation: the shoulder joint center lies beneath soft tissue and clothing, and small localization errors propagate into large angular errors for abduction, which references the vertical axis. All other metrics are robust to pitch type and delivery slot (overhand, three-quarter, sidearm), confirming that the validated formulas generalize across the mechanical diversity present in professional baseball.

\paragraph{Validation and scale.}
Each formula was derived iteratively: we computed the metric from pose data, compared against stadium tracking values on 13 professional pitchers (156 pitches, 8 right-throwing / 5 left-throwing, all 8 pitch types, overhand through sidearm deliveries, 75--102\,mph). We report MAE, maximum error, and Pearson $r$ for every metric in Table~\ref{tab:validation_summary}. A metric is considered \emph{validated} when MAE\,$<$\,1$^{\circ}$ (or $<$\,0.1\,ft) and $r > 0.95$. We apply the validated pipeline to 119,561 professional pitching sequences spanning eight pitch types. The resulting metric distributions align closely with normative values reported in large-scale biomechanical studies of professional and collegiate pitchers: lead-knee flexion at foot plant averages $\approx$45$^\circ$ (typically 35--55$^\circ$)~\cite{diffendaffer2023}, elbow flexion at maximum external rotation falls in the 100--120$^\circ$ range~\cite{diffendaffer2023,fleisig1999}, peak hip--shoulder separation reaches 40--60$^\circ$ in efficient deliveries~\cite{stodden2001}, and forward trunk tilt at ball release averages $\approx$35$^\circ$ (25--45$^\circ$)~\cite{diffendaffer2023}. No systematic outliers or drift are observed across pitch type or velocity band, confirming that the validated formulas transfer reliably to the broader professional population.

% \subsection{Application: Injury Risk Screening}

% As a downstream use case, we demonstrate automatic injury risk flagging using three validated metrics and clinically established thresholds~\cite{anz2010, fortenbaugh2009}: elbow flexion at MER above 100$^{\circ}$ (associated with elevated UCL stress), trunk lateral tilt at release above 30$^{\circ}$ (linked to shoulder impingement), and lead-knee flexion at foot plant below 30$^{\circ}$ (associated with increased ground reaction forces). A simple composite indicator,
% \begin{equation}
% \text{Risk} = w_1\,\mathbb{1}[\theta_{\text{elbow}}^{\text{MER}} > 100^{\circ}] + w_2\,\mathbb{1}[\phi_{\text{lateral}}^{\text{REL}} > 30^{\circ}] + w_3\,\mathbb{1}[\theta_{\text{knee}}^{\text{FP}} < 30^{\circ}],
% \end{equation}
% can be evaluated in real time from pose alone, enabling scalable screening without instrumented motion capture. We note that this formulation is illustrative; clinical deployment would require prospective validation against injury outcomes.

\begin{table*}[t]
\centering
\caption{Progression of injury-risk screening performance.}
\label{tab:injury_progression}
\begin{tabular}{lcccc}
\toprule
Version & Pitchers & TJ Cases (recent) & Features & Best AUC (TJ) \\
\midrule
V1 (static thresholds) & 7,348 & 209 & 5 & 0.503 \\
V2 (anomalies + workload) & 7,348 & 209 & ~25 & 0.633 \\
V3 (feature selection) & 7,348 & 209 & ~200 & 0.734 \\
V4 (full-scale) & 7,348 & 209 & ~70 & 0.776 \\
V5 (enriched + ensemble) & 7,348 & 209 & 144 & \textbf{0.811} \\
\bottomrule
\end{tabular}
\end{table*}

\begin{figure*}[t]
\centering
% \includesvg[width=0.9\linewidth]{Figures/feature.svg}
\includegraphics[width=0.9\linewidth]{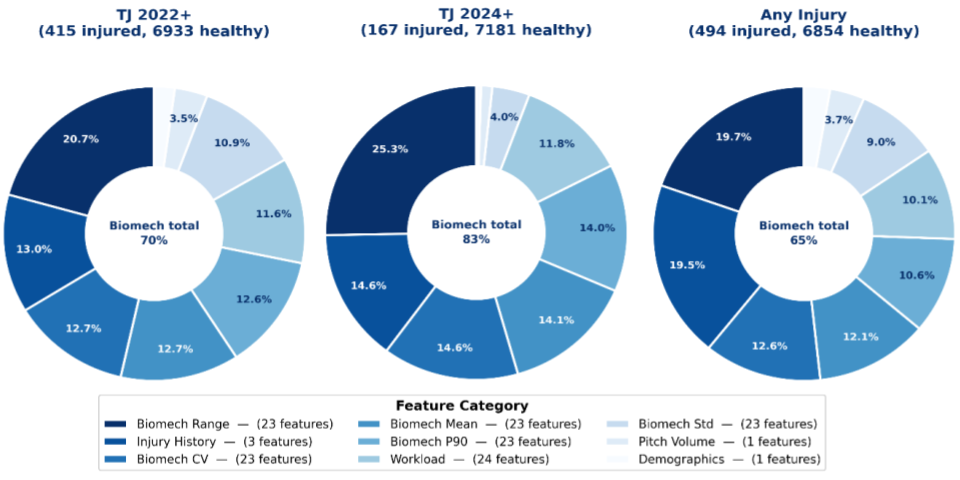}
\caption{\textbf{Feature Category Importance for Injury Prediction.} Distribution of GBM gain-based feature importance across nine feature categories for three injury targets}
\label{fig:feature_imp}
\end{figure*}

\subsection{Automated Injury-Risk Screening}

To demonstrate the clinical utility of our recovered biomechanical metrics, we develop an automated screening system that flags pitchers at elevated risk of Tommy John (UCL reconstruction) surgery or significant arm injuries. The system aggregates kinematic profiles from our monocular pipeline and combines them with workload, demographics, and medical history to produce actionable risk scores.

We iteratively scaled and refined the system across five versions, starting from simple literature-based thresholds on raw metrics (V1), progressing through personal anomaly detection (V1.5), workload integration (V2), large-scale feature selection (V3), and finally full-dataset aggregation with enriched statistics and ensemble modeling (V4--V5; see Table~\ref{tab:injury_progression} for progression details). The final ensemble, trained on 7,348 pitchers with 167 recent (2024+) Tommy John cases, achieves AUC 0.811 for Tommy John prediction and 0.825 for any significant arm injury, substantially outperforming earlier static-threshold (0.503) and anomaly-only (0.633) baselines.

\noindent \textbf{Feature importance insights.} GBM gain-based analysis reveals that biomechanical variability dominates kinematic signal: range and CV features together contribute 33--39\% of total importance across targets, consistently outranking mean values. P90 statistics are also highly predictive; for example, hip shoulder separation at foot plant ranks in the top 6--9 features across all targets, indicating that repeated exposure to extreme mechanics in the top 10\% of pitches is a stronger risk signal than average values alone. Medical history (prior TJ, prior injuries, IL years) accounts for 5--20\% despite comprising only three features, confirming past injury as the single strongest predictor. Pitch volume (\texttt{n\_pitches}) ranks top-3 across targets, reflecting cumulative stress exposure. Universally important biomechanics include range of elbow flexion at MER, shoulder rotation at release, trunk forward tilt at release, and knee flexion at release (Cohen's d 0.60--0.67)\footnote{Cohen's d measures the standardized difference between group means (injured vs.\ healthy pitchers); values 0.5--0.8 indicate medium-to-large effects.}, plus hip-shoulder separation P90 at foot plant, suggesting extreme or variable sequencing as a risk factor.

With such powerful injury-risk screening tools, coaches can threshold scores (e.g., top 10\% risk) to trigger workload adjustments, mechanical corrections, or medical evaluation, actions feasible at amateur, college, and training levels. The strong AUC on recent injuries demonstrates near-real-time relevance. These results validate that monocular 3D poses, when rigorously refined and aggregated with P90 and variability statistics, enable professional-grade injury-risk flagging without multi-camera infrastructure, closing the loop from affordable kinematic monitoring to preventive decision support.

%% file: sec/6_conc.tex
\section{Discussion}
\label{sec:discussion}

Three factors enable tracking-consistent measurements from monocular video. First, PitcherNet's 1.8\,mm mean joint error provides a sufficiently accurate geometric foundation: at typical limb lengths, this translates to sub-degree angular noise, well below our 1$^{\circ}$ validation threshold for most metrics. Second, our iterative validation protocol, deriving each formula against a professional tracking reference rather than assuming correctness from textbook geometry, catches subtle mismatches that would otherwise introduce systematic bias. Third, using relative angles (joint-to-joint) rather than absolute positions wherever possible makes the formulas largely viewpoint-invariant.

The principal limitation is shoulder abduction, where MAE reaches 6.2$^{\circ}$ for the throwing arm and 21.4$^{\circ}$ for the glove side. This is not a formula deficiency but a fundamental challenge of markerless pose estimation: the glenohumeral joint center lies beneath soft tissue and clothing, and small localization errors propagate into large angular deviations when the reference axis is vertical. All other validated metrics are robust to pitch type, delivery slot, and velocity band, confirming generalization across the mechanical diversity in professional baseball. 
% More broadly, our pipeline inherits whatever accuracy the upstream pose estimator provides. Motion blur, partial occlusion, and low broadcast resolution degrade pose quality and, consequently, metric accuracy, though we have not quantified this propagation in a controlled ablation. Monocular depth ambiguity is a further source of error for metrics that depend on the $y$-axis (depth toward home plate); multi-view input or explicit depth priors could reduce this.

Compared to laboratory marker-based systems (Vicon, Qualisys), which achieve $<$\,0.1$^{\circ}$ MAE, our approach trades a small accuracy margin (0.5$^{\circ}$ typical vs.\ 0.1$^{\circ}$) for markerless operation, field deployment on broadcast footage, zero hardware cost beyond a camera, and fully automated processing. For most coaching and screening applications, the sub-degree accuracy of our validated metrics is more than sufficient. 
% The methodology also generalizes beyond baseball: any overhead throwing sport (cricket bowling, javelin, tennis serve) or rotational sport (golf) involves analogous kinematic chains, and the same validation protocol, iterate against a sport-specific gold standard, can be applied to derive and verify the corresponding formulas.

Several directions remain open. Multi-view integration could improve depth accuracy and shoulder localization. Temporal modeling (angular velocities, acceleration profiles) would extend the current per-frame metrics into continuous biomechanical time series. Pitcher-specific baselines would enable anomaly detection relative to an individual's own norms rather than population averages. Finally, prospective validation of the injury-risk screening application against actual injury outcomes is needed before clinical deployment.

\section{Conclusion}
\label{sec:conclusion}

We introduced a monocular pipeline that takes broadcast video, estimates 3D pose, and applies a refinement stack (global lifting, bone-length enforcement, constrained IK, smoothing, and symmetry) before computing 18 clinically relevant pitching metrics validated against professional tracking data. Across 13 professional pitchers, 16 of 18 metrics meet the sub-degree threshold, and the same formulas scale to 119,561 sequences without drift across pitch type or velocity band. We then operationalize these metrics for injury prevention: an automated screening model that aggregates kinematics with workload and medical history achieves AUC 0.811 for Tommy John prediction and 0.825 for significant arm injuries on 7,348 pitchers, showing that the signals are strong enough for practical risk stratification. In practice, the system supports scalable monitoring, periodic screening, and targeted workload or mechanics interventions in environments where multi-camera motion capture is infeasible.

%% file: main.bib
@inproceedings{bright2024pitchernet,
  author    = {Bright, Jerrin and Balaji, Balaji and Chen, Yimu and Clausi, David A. and Zelek, John S.},
  title     = {PitcherNet: Powering the Moneyball Evolution in Baseball Video Analytics},
  booktitle = {Proceedings of the IEEE/CVF Conference on Computer Vision and Pattern Recognition Workshops (CVPRW)},
  year      = {2024},
  pages     = {769--787}
}

@article{li2021mhformer,
  author  = {Li, Wenhao and Liu, Hong and Tang, Hao and Wang, Peng and Van Gool, Luc},
  title   = {MHFormer: Multi-Hypothesis Transformer for 3D Human Pose Estimation},
  journal = {arXiv preprint arXiv:2111.12707},
  year    = {2021}
}

@article{aristidou2011fabrik,
  title={FABRIK: A fast, iterative solver for the Inverse Kinematics problem},
  author={Aristidou, Andreas and Lasenby, Joan},
  journal={Graphical Models},
  volume={73},
  number={5},
  pages={243--260},
  year={2011},
  publisher={Elsevier}
}

@article{fortenbaugh2009,
  title={Baseball pitching biomechanics in relation to injury risk and performance},
  author={Fortenbaugh, Dave and Fleisig, Glenn S and Andrews, James R},
  journal={Sports Health},
  volume={1},
  number={4},
  pages={314--320},
  year={2009}
}

@article{anz2010,
  title={Correlation of torque and elbow injury in professional baseball pitchers},
  author={Anz, Adam W and Bushnell, Brandon D and Griffin, Leanne P and Noonan, Thomas J and Torry, Michael R and Hawkins, Richard J},
  journal={The American journal of sports medicine},
  volume={38},
  number={7},
  pages={1368--1374},
  year={2010}
}

@article{stodden2001,
  title={Relationship of pelvis and upper torso kinematics to pitched baseball velocity},
  author={Stodden, David F and Fleisig, Glenn S and McLean, Scott P and Lyman, Sherry L and Andrews, James R},
  journal={Journal of applied biomechanics},
  volume={17},
  number={2},
  pages={164--172},
  year={2001}
}

@article{kanazawa2018,
  title={End-to-end recovery of human shape and pose},
  author={Kanazawa, Angjoo and Black, Michael J and Jacobs, David W and Malik, Jitendra},
  journal={Proceedings of the IEEE conference on computer vision and pattern recognition},
  pages={7122--7131},
  year={2018}
}

@article{bright2025dream,
  title={Dreampose3d: Hallucinative diffusion with prompt learning for 3d human pose estimation},
  author={Bright, Jerrin and Chen, Yuhao and Zelek, John S},
  journal={arXiv preprint arXiv:2511.09502},
  year={2025}
}

@inproceedings{bright2023mitigating,
  title={Mitigating motion blur for robust 3D baseball player pose modeling for pitch analysis},
  author={Bright, Jerrin and Chen, Yuhao and Zelek, John},
  booktitle={Proceedings of the 6th International Workshop on Multimedia Content Analysis in Sports},
  pages={63--71},
  year={2023}
}

@article{oyama2012baseball,
  title={Baseball pitching kinematics, joint loads, and injury prevention},
  author={Oyama, Sakiko},
  journal={Journal of sport and health science},
  volume={1},
  number={2},
  pages={80--91},
  year={2012},
  publisher={Elsevier}
}

@article{erickson2016predicting,
  title={Predicting and preventing injury in Major League Baseball},
  author={Erickson, Brandon J and Chalmers, Peter N and Bush-Joseph, Charles A and Romeo, Anthony A},
  journal={Am J Orthop},
  volume={45},
  number={3},
  pages={152--156},
  year={2016}
}

@article{ramirez2016two,
  title={Two-Dimensional Video Analysis of Youth and Adolescent Pitching Biomechanics: A Tool For the Common Athlete},
  author={Ramirez, Victor and others},
  journal={Current Sports Medicine Reports},
  volume={15},
  number={5},
  pages={336--343},
  year={2016},
  doi={10.1249/JSR.0000000000000295},
}

@article{defroda2016two,
  title={Two-dimensional video analysis of youth and adolescent pitching biomechanics: A tool for the common athlete},
  author={DeFroda, Steven F and Thigpen, Charles A and Kriz, Peter K},
  journal={Current Sports Medicine Reports},
  volume={15},
  number={5},
  pages={350--358},
  year={2016},
  publisher={LWW}
}

@article{singh2012hawk,
  title={Hawk Eye: A Logical Innovative Technology Use in Sports for Effective Decision Making.},
  author={Singh Bal, Baljinder and Dureja, Gaurav},
  journal={Sport Science Review},
  volume={21},
  year={2012}
}

@inproceedings{li2021baseball,
  title={Baseball swing pose estimation using openpose},
  author={Li, Yung-Che and Chang, Ching-Tang and Cheng, Chin-Chang and Huang, Yu-Len},
  booktitle={2021 IEEE International Conference on Robotics, Automation and Artificial Intelligence (RAAI)},
  pages={6--9},
  year={2021},
  organization={IEEE}
}

@article{chiu20243d,
  title={3D baseball pitcher pose reconstruction using joint-wise volumetric triangulation and baseball customized filter system},
  author={Chiu, Yun-Wei and Huang, Kuei-Ting and Wu, Yuh-Renn and Huang, Jyh-How and Hsu, Wei-Li and Wu, Pei-Yuan},
  journal={IEEE Access},
  volume={12},
  pages={117110--117125},
  year={2024},
  publisher={IEEE}
}

@article{diffendaffer2023,
  title   = {The Clinician's Guide to Baseball Pitching Biomechanics},
  author  = {Diffendaffer, Alek Z and Bagwell, Michael S and Fleisig, Glenn S and others},
  journal = {Sports Health},
  volume  = {15},
  number  = {2},
  pages   = {164--172},
  year    = {2023}
}

@article{fleisig1999,
  title   = {Kinematic and kinetic comparison of baseball pitching among various levels of development},
  author  = {Fleisig, Glenn S and others},
  journal = {Journal of Biomechanics},
  year    = {1999}
}

@article{mendes2023survey,
  title={A survey of advanced computer vision techniques for sports},
  author={Mendes-Neves, Tiago and Meireles, Lu{\~A}s and Mendes-Moreira, Jo{\~A} and others},
  journal={arXiv preprint arXiv:2301.07583},
  year={2023}
}

@article{vats2023player,
  title={Player tracking and identification in ice hockey},
  author={Vats, Kanav and Walters, Pascale and Fani, Mehrnaz and Clausi, David A and Zelek, John S},
  journal={Expert systems with applications},
  volume={213},
  pages={119250},
  year={2023},
  publisher={Elsevier}
}

@inproceedings{sun2024gta,
  title={Gta: Global tracklet association for multi-object tracking in sports},
  author={Sun, Jiacheng and Huang, Hsiang-Wei and Yang, Cheng-Yen and Jiang, Zhongyu and Hwang, Jenq-Neng},
  booktitle={Proceedings of the Asian Conference on Computer Vision},
  pages={421--434},
  year={2024}
}

@inproceedings{yamada2025trackid3x3,
  title={TrackID3x3: A Dataset and Algorithm for Multi-Player Tracking with Identification and Pose Estimation in 3x3 Basketball Full-court Videos},
  author={Yamada, Kazuhiro and Yin, Li and Hu, Qingrui and Ding, Ning and Iwashita, Shunsuke and Ichikawa, Jun and Kotani, Kiwamu and Yeung, Calvin and Fujii, Keisuke},
  booktitle={Proceedings of the 8th International ACM Workshop on Multimedia Content Analysis in Sports},
  pages={163--173},
  year={2025}
}

@article{ghasemzadeh2021deepsportlab,
  title={Deepsportlab: a unified framework for ball detection, player instance segmentation and pose estimation in team sports scenes},
  author={Ghasemzadeh, Seyed Abolfazl and Van Zandycke, Gabriel and Istasse, Maxime and Sayez, Niels and Moshtaghpour, Amirafshar and De Vleeschouwer, Christophe},
  journal={arXiv preprint arXiv:2112.00627},
  year={2021}
}

@article{bock2015pitch,
  title={Pitch sequence complexity and long-term pitcher performance},
  author={Bock, Joel R},
  journal={Sports},
  volume={3},
  number={1},
  pages={40--55},
  year={2015},
  publisher={MDPI}
}

@article{sidle2018using,
  title={Using multi-class classification methods to predict baseball pitch types},
  author={Sidle, Glenn and Tran, Hien},
  journal={Journal of Sports Analytics},
  volume={4},
  number={1},
  pages={85--93},
  year={2018},
  publisher={SAGE Publications Sage UK: London, England}
}

@article{hickey2020dissecting,
  title={Dissecting moneyball: Improving classification model interpretability in baseball pitch prediction},
  author={Hickey, Kevin and Zhou, Lina and Tao, Jie},
  year={2020}
}

@article{oyama2017reliability,
  title={Reliability and validity of quantitative video analysis of baseball pitching motion},
  author={Oyama, Sakiko and Sosa, Araceli and Campbell, Rebekah and Correa, Alexandra},
  journal={Journal of applied biomechanics},
  volume={33},
  number={1},
  pages={64--68},
  year={2017},
  publisher={Human Kinetics, Inc.}
}

@article{oeding2024pitch,
  title={Pitch-tracking metrics as a predictor of future shoulder and elbow injuries in major league baseball pitchers: a machine-learning and game-theory based analysis},
  author={Oeding, Jacob F and Boos, Alexander M and Kalk, Josh R and Sorenson, Dane and Verhooven, F Martijn and Moatshe, Gilbert and Camp, Christopher L},
  journal={Orthopaedic Journal of Sports Medicine},
  volume={12},
  number={8},
  pages={23259671241264260},
  year={2024},
  publisher={SAGE Publications Sage CA: Los Angeles, CA}
}

@article{karnuta2020machine,
  title={Machine learning outperforms regression analysis to predict next-season Major League Baseball player injuries: epidemiology and validation of 13,982 player-years from performance and injury profile trends, 2000-2017},
  author={Karnuta, Jaret M and Luu, Bryan C and Haeberle, Heather S and Saluan, Paul M and Frangiamore, Salvatore J and Stearns, Kim L and Farrow, Lutul D and Nwachukwu, Benedict U and Verma, Nikhil N and Makhni, Eric C and others},
  journal={Orthopaedic journal of sports medicine},
  volume={8},
  number={11},
  pages={2325967120963046},
  year={2020},
  publisher={SAGE Publications Sage CA: Los Angeles, CA}
}

@article{kang2025data,
  title={Data-driven approaches for predicting tommy john surgery risk in major league baseball pitchers},
  author={Kang, Bosuk and Park, Minsu and del Pobil, Angel P and Park, Eunil},
  journal={Journal of Big Data},
  volume={12},
  number={1},
  pages={87},
  year={2025},
  publisher={Springer}
}

@inproceedings{piergiovanni2019early,
  title={Early detection of injuries in mlb pitchers from video},
  author={Piergiovanni, AJ and Ryoo, Michael S},
  booktitle={Proceedings of the IEEE/CVF conference on computer vision and pattern recognition workshops},
  pages={0--0},
  year={2019}
}

@article{sutter2018predicting,
  title={Predicting injury in professional baseball pitchers from delivery mechanics: a statistical model using quantitative video analysis},
  author={Sutter, E Grant and Orenduff, Justin and Fox, Will J and Myers, Joshua and Garrigues, Grant E},
  journal={Orthopedics},
  volume={41},
  number={1},
  pages={43--53},
  year={2018},
  publisher={SLACK Incorporated Thorofare, NJ}
}

@article{bullock2021baseball,
  title={Baseball pitching biomechanics in relation to pain, injury, and surgery: A systematic review},
  author={Bullock, Garrett S and Menon, Gautam and Nicholson, Kristen and Butler, Robert J and Arden, Nigel K and Filbay, Stephanie R},
  journal={Journal of science and medicine in sport},
  volume={24},
  number={1},
  pages={13--20},
  year={2021},
  publisher={Elsevier}
}

@article{moore2025reverse,
  title={Reverse (Bio) Engineering: A Machine Learning Approach to Optimize Baseball Pitcher Health and Performance},
  author={Moore, Robert C},
  year={2025}
}

@article{solomito2024lead,
  title={Lead knee flexion angle is associated with both ball velocity and upper extremity joint moments in collegiate baseball pitchers},
  author={Solomito, Matthew J and Garibay, Erin J and Cohen, Andrew and Nissen, Carl W},
  journal={Sports Biomechanics},
  volume={23},
  number={12},
  pages={2626--2636},
  year={2024},
  publisher={Taylor \& Francis}
}

@article{gauthier2025evaluation,
  title={Evaluation and Treatment of Baseball Pitchers: There’s More to Assess than the Arm},
  author={Gauthier, Matthew L and Unverzagt, Casey A and Davies, George J},
  journal={International journal of sports physical therapy},
  volume={20},
  number={1},
  pages={113},
  year={2025}
}

@article{manzi2022increased,
  title={Increased elbow and olecranon injury history in professional pitchers with increased elbow flexion at ball release},
  author={Manzi, Joseph E and Ciccotti, Michael C and Trauger, Nicolas and Black, Grant G and Thacher, Ryan R and Boddapati, Venkat and Dines, Joshua S},
  journal={The American Journal of Sports Medicine},
  volume={50},
  number={4},
  pages={1054--1060},
  year={2022},
  publisher={SAGE Publications Sage CA: Los Angeles, CA}
}

@article{solomito2015lateral,
  title={Lateral trunk lean in pitchers affects both ball velocity and upper extremity joint moments},
  author={Solomito, Matthew J and Garibay, Erin J and Woods, Jessica R and {\~O}unpuu, Sylvia and Nissen, Carl W},
  journal={The American journal of sports medicine},
  volume={43},
  number={5},
  pages={1235--1240},
  year={2015},
  publisher={SAGE Publications Sage CA: Los Angeles, CA}
}

@article{peters2025association,
  title={Association between pitching velocity and elbow varus torque},
  author={Peters, Scott and Bullock, Garrett S and Nicholson, Kristen F},
  journal={Brazilian Journal of Physical Therapy},
  volume={29},
  number={5},
  pages={101222},
  year={2025},
  publisher={Elsevier}
}

@article{camp2017relationship,
  title={The relationship of throwing arm mechanics and elbow varus torque: within-subject variation for professional baseball pitchers across 82,000 throws},
  author={Camp, Christopher L and Tubbs, Travis G and Fleisig, Glenn S and Dines, Joshua S and Dines, David M and Altchek, David W and Dowling, Brittany},
  journal={The American journal of sports medicine},
  volume={45},
  number={13},
  pages={3030--3035},
  year={2017},
  publisher={SAGE Publications Sage CA: Los Angeles, CA}
}

@article{gran2024deep,
  title={Deep HM-SORT: Enhancing Multi-Object Tracking in Sports with Deep Features, Harmonic Mean, and Expansion IOU},
  author={Gran-Henriksen, Matias and Lindgaard, Hans Andreas and Kiss, Gabriel and Lindseth, Frank},
  journal={arXiv preprint arXiv:2406.12081},
  year={2024}
}
